\documentclass[conference]{IEEEtran}
\IEEEoverridecommandlockouts
\usepackage{cite}
\usepackage{amsmath,amssymb,amsfonts}
\usepackage{algorithm}
\usepackage{algpseudocode}
\usepackage{graphicx}
\usepackage{textcomp}
\usepackage{xcolor}
\usepackage{hyperref}
\hypersetup{
    colorlinks=true,
    linkcolor=blue,
    filecolor=magenta,      
    urlcolor=cyan,
    pdftitle={Overleaf Example},
    pdfpagemode=FullScreen,
    }
\def\BibTeX{{\rm B\kern-.05em{\sc i\kern-.025em b}\kern-.08em
    T\kern-.1667em\lower.7ex\hbox{E}\kern-.125emX}}
\begin{document}

\title{Energy Prediction using Federated Learning}

\author{\IEEEauthorblockN{Meghana Bharadwaj}
\IEEEauthorblockA{\textit{Department of Electrical Engineering} \\
\textit{Stanford University}\\
mbhar@stanford.edu}
\and
\IEEEauthorblockN{Sanjana Sarda}
\IEEEauthorblockA{\textit{Department of Electrical Engineering} \\
\textit{Stanford University}\\
ssarda@stanford.edu}
}

\maketitle

\begin{abstract}
%notdone
In this work, we demonstrate the viability of using federated learning to successfully predict energy consumption as well as solar production for all households within a certain network using low-power and low-space consuming embedded devices. We also demonstrate our prediction performance improving over time without the need for sharing private consumer energy data. We simulate a system with 4 nodes with 1 year's worth of data to show this.
\end{abstract}

% \begin{IEEEkeywords}
% component, formatting, style, styling, insert
% \end{IEEEkeywords}

\section{Introduction}
%notdone
As households have more distributed energy devices such as electric vehicles, solar panels, and residential batteries, it is necessary to operate them optimally to save costs and reduce energy consumption. With improved energy usage prediction and solar forecasting, it is also possible to participate effectively in demand response programs. Energy usage patterns of households can be vastly different, therefore it is beneficial to customize the energy management prediction and control strategy per household. Furthermore, it is increasingly common for households to have smart devices and other load-controlling technologies however most of these do not have on-device machine learning capabilities so this project aims to introduce more intelligent predictions through federated learning. 

This makes it a suitable application for distributed machine learning. Due to privacy and latency reasons, it is beneficial to train and run the energy prediction model locally on an edge device. We want the device to be able to adapt its prediction in real-time based on variations in weather conditions and usage patterns for each user while improving the overall global model. Through federated learning, the global model is able to learn from a wider range of data, from all of the nodes in the network, without compromising user privacy. 

\section{Related Work}
The authors of \cite{ev_pred} apply distributed machine learning to predict the energy demand of EV charging stations. They use a Deep Neural Net model with approximately 65,000 parameters. 

Tun et al \cite{fedlearn_energypred} use federated learning for energy demand prediction. They use a bidirectional LSTM model where the input features are the previous 30 days' worth of energy demand data. Since they have more nodes in their network, they use clustering to first aggregate nodes with similar energy usage patterns and then perform federated learning on clusters instead of individual nodes. This will not be necessary for our system since we only have 4 nodes. 

In \cite{elec_load_forecast}, the authors use an LSTM-based model for energy prediction. They perform updates on batches of clients instead of using all of the available nodes for each federated learning update. The model is trained and tested on household electricity consumption data from the Pecan Street database.

\section{Problem Statement}
%notdone
We propose using distributed machine learning to predict household energy consumption and solar prediction for a group of houses via Federated Learning.
\begin{figure}[ht]
    \centering
    \includegraphics[width=0.4\textwidth]{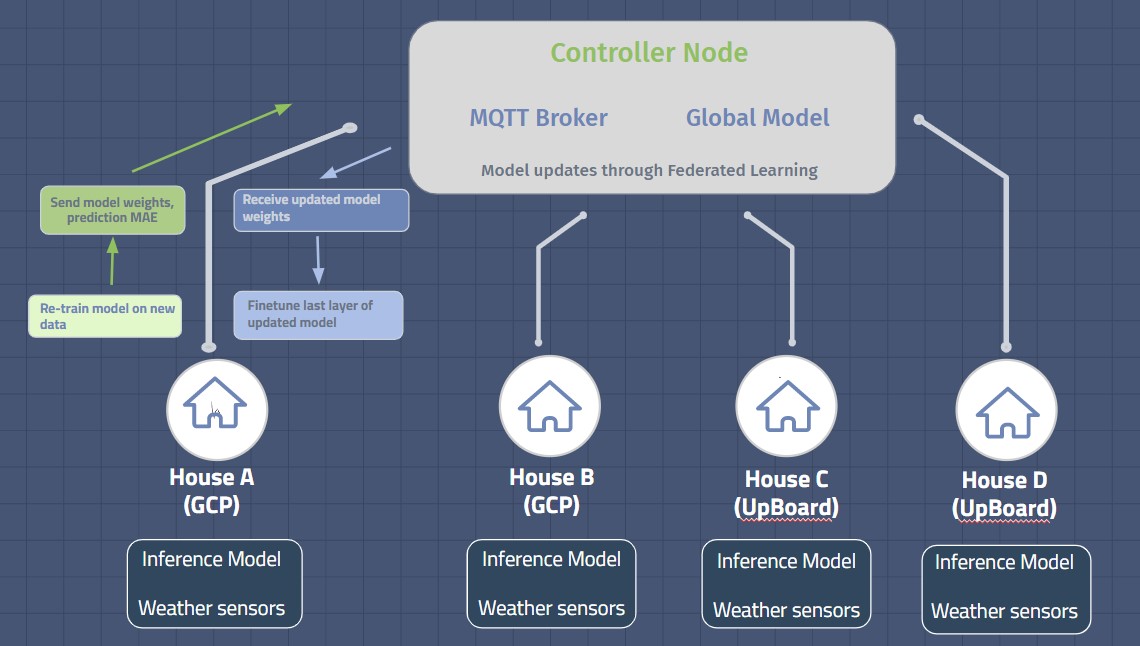}
    \caption{System Diagram}
    \label{fig:sysd}
\end{figure}

\section{Data}
The data used to train and test our model and federated learning algorithms is sourced from the Pecan Street database. We used household electricity consumption and solar power production data recorded on a 15-minute time resolution. However since we are doing hourly predictions, we pre-processed the raw data to obtain aggregated hourly energy and solar output values. Our dataset consists of 20 households located in Austin, Texas. All of the households have solar panels. We also used the weather data available from Pecan Street which contains humidity, temperature, cloud cover, wind speed, dew point, and precipitation intensity. The weather data was available for 4 locations in Texas so we chose the closest location to Austin. The data spans 1 year. 

\subsection{Train Data}
The base model is trained using 16 houses' worth of data as well as half of the available data for 4 households. The input features are the previous 30 days' worth of energy consumption and solar prediction values, time, and weather parameters at the current time step. 

\subsection{Test Data}
The federated learning algorithm is tested on 6 months worth of data on 4 households. 

\section{Machine Learning}
The machine learning model is responsible for predicting hourly energy consumption and solar power production. We tested both 1-hour ahead predictions as well as 6-hour ahead predictions. 

\subsection{Model Architecture}
The base model is a 4 layer neural network with approximately 400,000 parameters. We used dense layers with Relu activation.
The model was trained on the dataset for 9 epochs with a learning rate of 0.001. 

\subsection{Federated Learning}
We use the FedAvg algorithm to update the global model based on the individually retrained models at each node. For all 4 layers of the neural network, the following update is performed:

\begin{equation}
f(w) = \sum^K_{k=1} \frac{n_k}{n} F_k(w)
\end{equation}

where K is the total number of nodes in the network, $n_k$ is the number of data points that each local model is trained on, $n$ is the total number of data points across all nodes, $F_k(w)$ are the weights of each local model and $f(w)$ are the global model weights. 

\begin{table*}[t]
  \centering
  \begin{tabular}{l | l l l l }
    \textbf{Training Phase} & \textbf{Layers Trained} & \textbf{Epochs} & \textbf{Batch Size}&\textbf{Learning Rate}\\
    Initial Model & All & 5 & 32 & 0.001  \\
    Retrain Model with New Data & All & 2 & 32 & 0.00001  \\
    Fine-tune Global Model with New Data & Last & 2 & 32 & 0.01 \\
  \end{tabular}
  \vspace{3mm}
  \caption{Training Details}
  \label{tab:perf}
\end{table*}

\section{Implementation}
%notdone
\subsection{Hardware Setup}
In a real system, each household in the area of interest would be a node in the network. Since we are limited by the number of nodes available to us, we are simulating this network using 2 upboards and 2 GCE worker nodes. While this setup worked, we expect that we can replace the upboards with raspberry pi zeros in future iterations. 

Household energy consumption and solar production are dependent on weather parameters such as temperature, humidity, and irradiation. Each node will have weather sensors that provide this input data to the prediction model.

It is not possible to test on physical household and EV charging/solar panel hardware so we do this through simulation. However, in order to simulate data processing on each device, we work with raw data and add an initial scaling step to our local models. 

\subsection{Workflow}
Each worker node is set up with a trained neural network that it uses to run hourly inference on local data. This is unique to each household. The node additionally collects two weeks' worth of local data and uses this to train its individual model. The re-training is done with a low learning rate of 1e-5. Provided that the performance of the trained model improves, each node will send this over to the global controller node. 

The controller node waits until it has received each worker node's model within a defined network, after which it uses FedAvg to compute the new overall model. Models from each worker node are equally weighted as they are trained on the same number of data samples. Once this new model is produced, the controller node sends this over to each of the nodes within the network. 

After the worker node sends over its trained model to the controller node, it waits to receive the updated overall model from the controller node before proceeding with further inference. To ensure accuracy on local data for the sake of customization, the worker node fine-tunes the last layer of the updated overall model with the two weeks worth of new data. Fine-tuning is done with a learning rate of 1e-2. This process will then continue with the fine-tuned model. 

\subsection{Communication}
Our system uses MQTT (python paho-mqtt) for worker and controller communication with the Mosquitto broker located on the controller node. Weights are serialized into byte objects using Pickle during publishing and then deserialized on subscription. To ensure receipt of the updated model from the controller node, we use an MQTT loop on the worker node that keeps the subscriber alive until the model has been received. 

\subsection{Deployment}
The original trained neural net model along with all requirements and scripts are packaged within a unique docker container that is then accessed by each worker node. The controller node uses K3s to run the inference script within the docker container located on the worker node. Each household worker node receives a unique label for identification. This label is also used to correctly start a pod on a specific worker node. 

\section{Results}
We tested our system through simulation, using 6 months of data for each of the 4 households in our network. The simulation is conducted for both 1-hour ahead predictions as well as 6-hour ahead predictions. 

The following graphs show the predicted energy consumption and solar power output values and the corresponding mean absolute error. 

\begin{figure}[!h]
    \centering
    \includegraphics[width=0.4\textwidth]{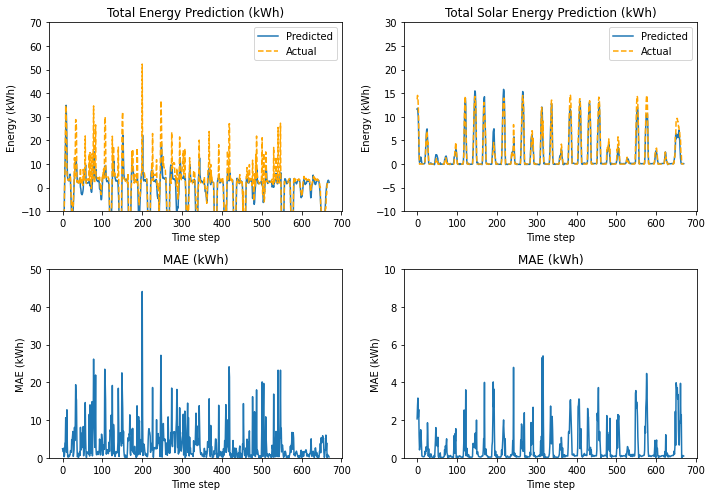}
    \caption{House A: Predictions}
    \label{fig:preda}
\end{figure}

\begin{figure}[!h]
    \centering
    \includegraphics[width=0.4\textwidth]{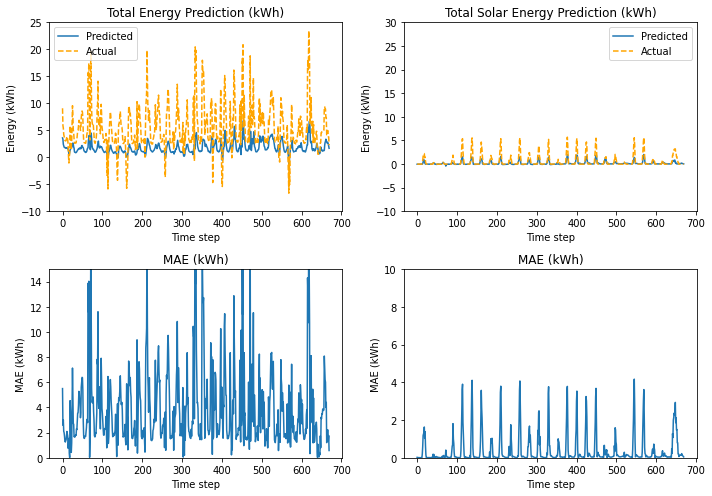}
    \caption{House B: Predictions}
    \label{fig:predb}
\end{figure}

\begin{figure}[!h]
    \centering
    \includegraphics[width=0.4\textwidth]{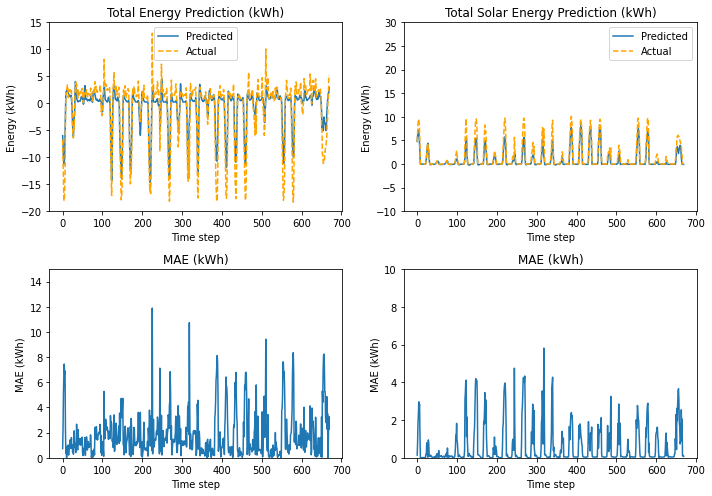}
    \caption{House C: Predictions}
    \label{fig:predc}
\end{figure}

\begin{figure}[!h]
    \centering
    \includegraphics[width=0.4\textwidth]{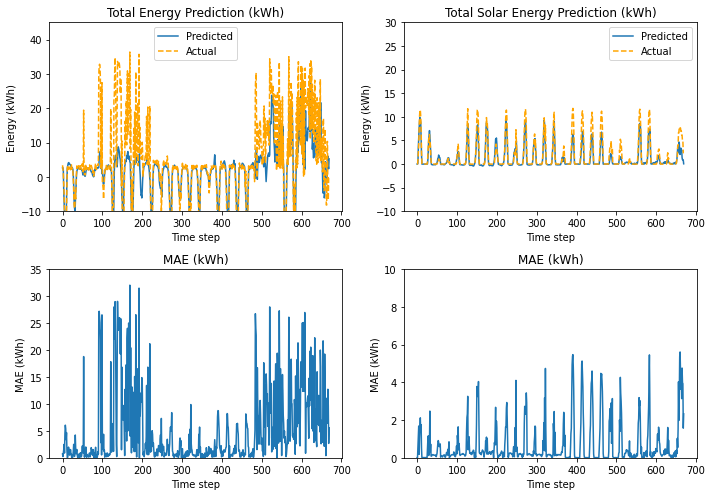}
    \caption{House D: Predictions}
    \label{fig:predd}
\end{figure}

See Figure \ref{fig:preda}, Figure \ref{fig:predb}, Figure \ref{fig:predc}, and Figure \ref{fig:predd} for the plots of all 4 houses for 1-hour ahead predictions.

\begin{figure}[!h]
    \centering
    \includegraphics[width=0.4\textwidth]{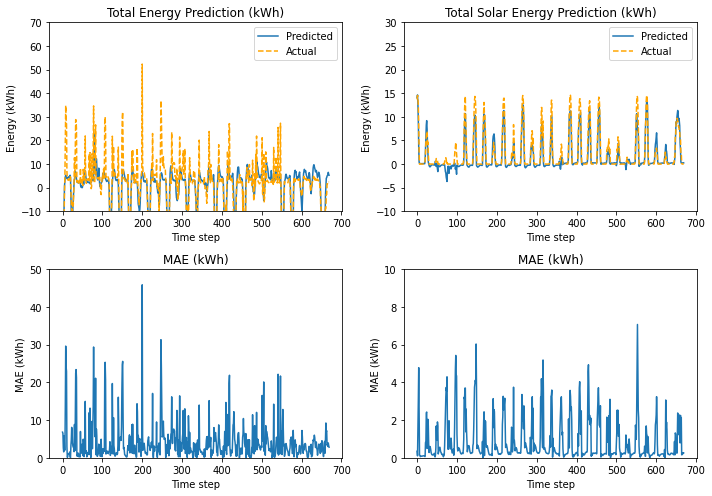}
    \caption{House A: Predictions}
    \label{fig:predas}
\end{figure}

\begin{figure}[!h]
    \centering
    \includegraphics[width=0.4\textwidth]{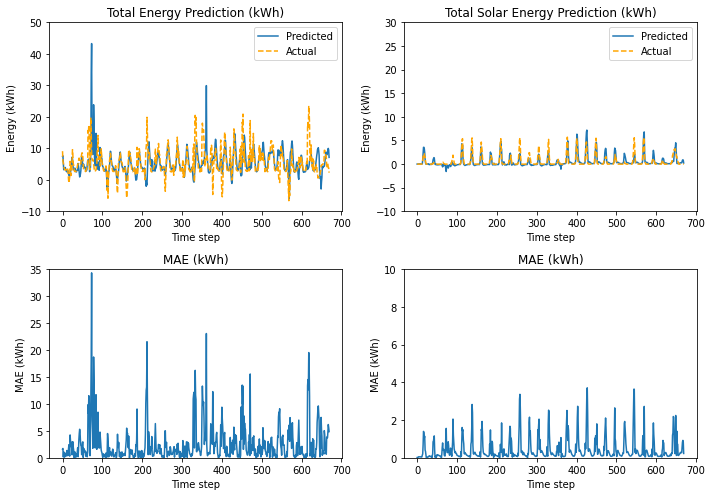}
    \caption{House B: Predictions}
    \label{fig:predbs}
\end{figure}

\begin{figure}[!h]
    \centering
    \includegraphics[width=0.4\textwidth]{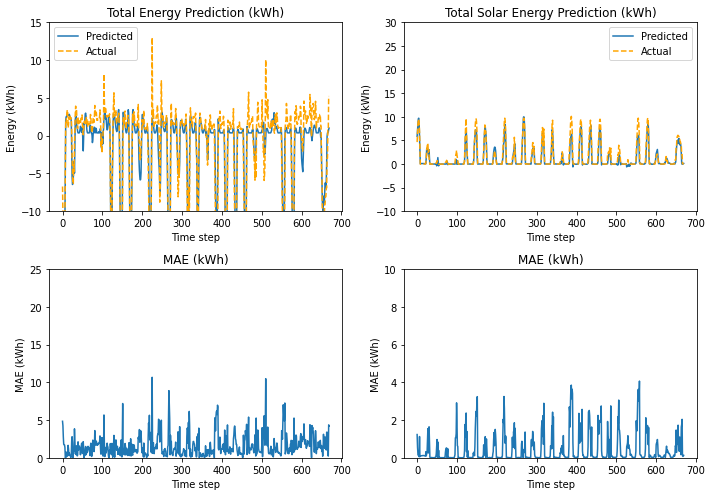}
    \caption{House C: Predictions}
    \label{fig:predcs}
\end{figure}

\begin{figure}[!h]
    \centering
    \includegraphics[width=0.4\textwidth]{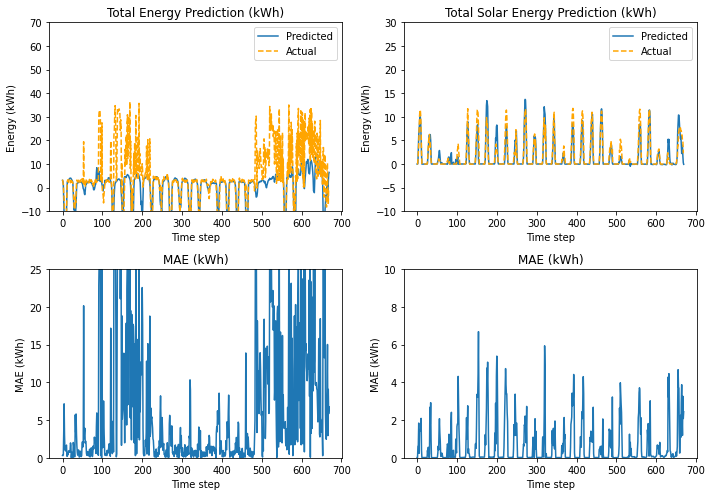}
    \caption{House D: Predictions}
    \label{fig:predds}
\end{figure}

See Figure \ref{fig:predas}, Figure \ref{fig:predbs}, Figure \ref{fig:predcs}, and Figure \ref{fig:predds} for the plots of all 4 houses for 6-hour ahead predictions.

As can be seen, the solar predictions have low MAE since they follow a regular pattern. The energy predictions have higher MAE, in particular when there are large spikes in consumption as the model is not able to predict these extreme values accurately. 

The following graphs show the mean absolute error of every 2 weeks of data before and after federated learning updates are made. For all nodes, for both 1-hour and 6-hour ahead predictions, the MAE after federated learning updates are performed is lower than the MAE calculated with the previous model. This shows that federated learning is helping to improve the accuracy of the model. Furthermore, we see that the MAE decreases throughout the simulation showing that the models are improving in accuracy overall. 

\begin{figure}[!h]
    \centering
    \includegraphics[width=0.4\textwidth]{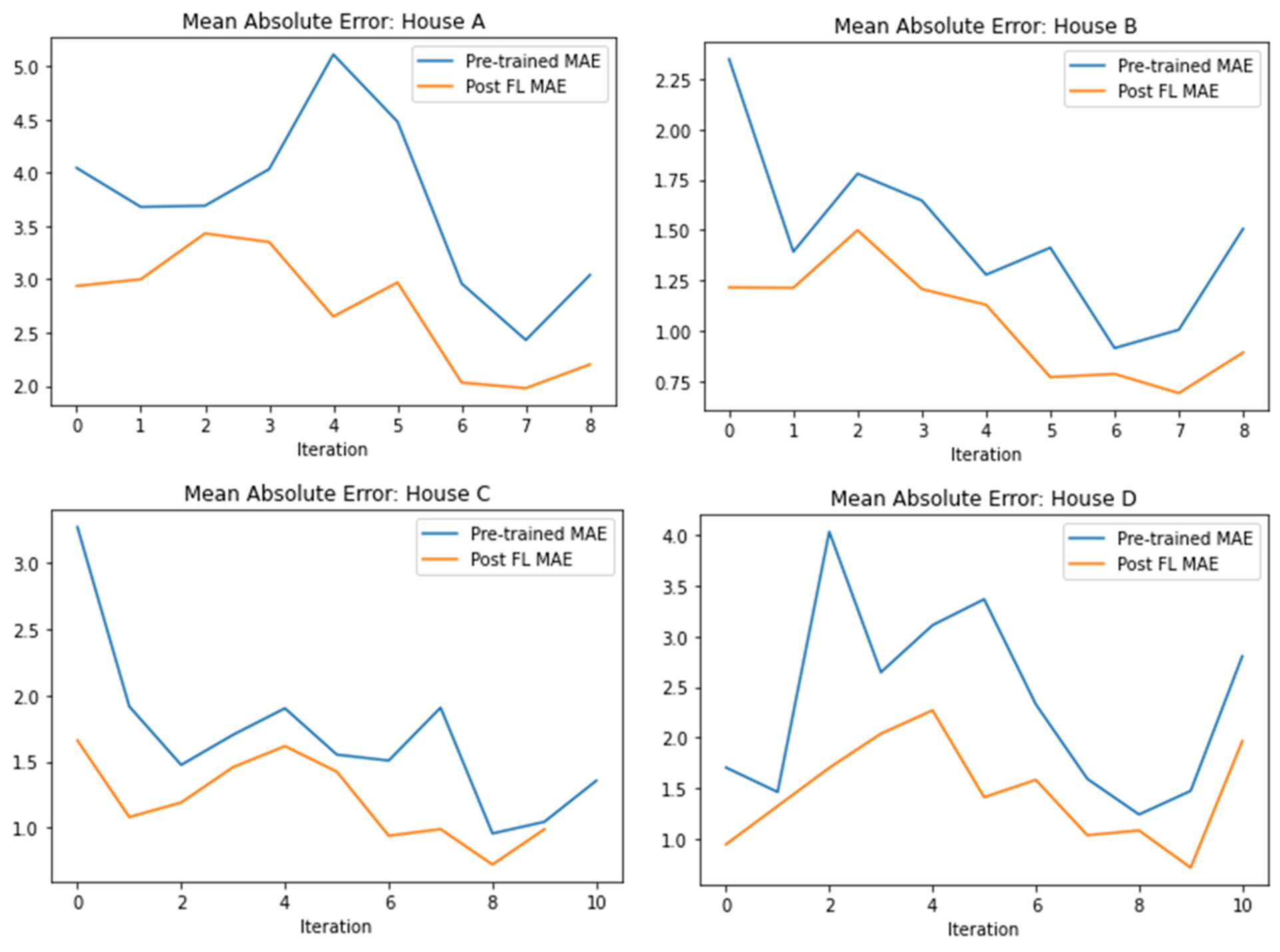}
    \caption{MAE per 2-week iteration for all 4 houses (1-hr ahead)}
    \label{fig:maeone}
\end{figure}

1-hour ahead predictions: Figure \ref{fig:maeone}

\begin{figure}[!h]
    \centering
    \includegraphics[width=0.4\textwidth]{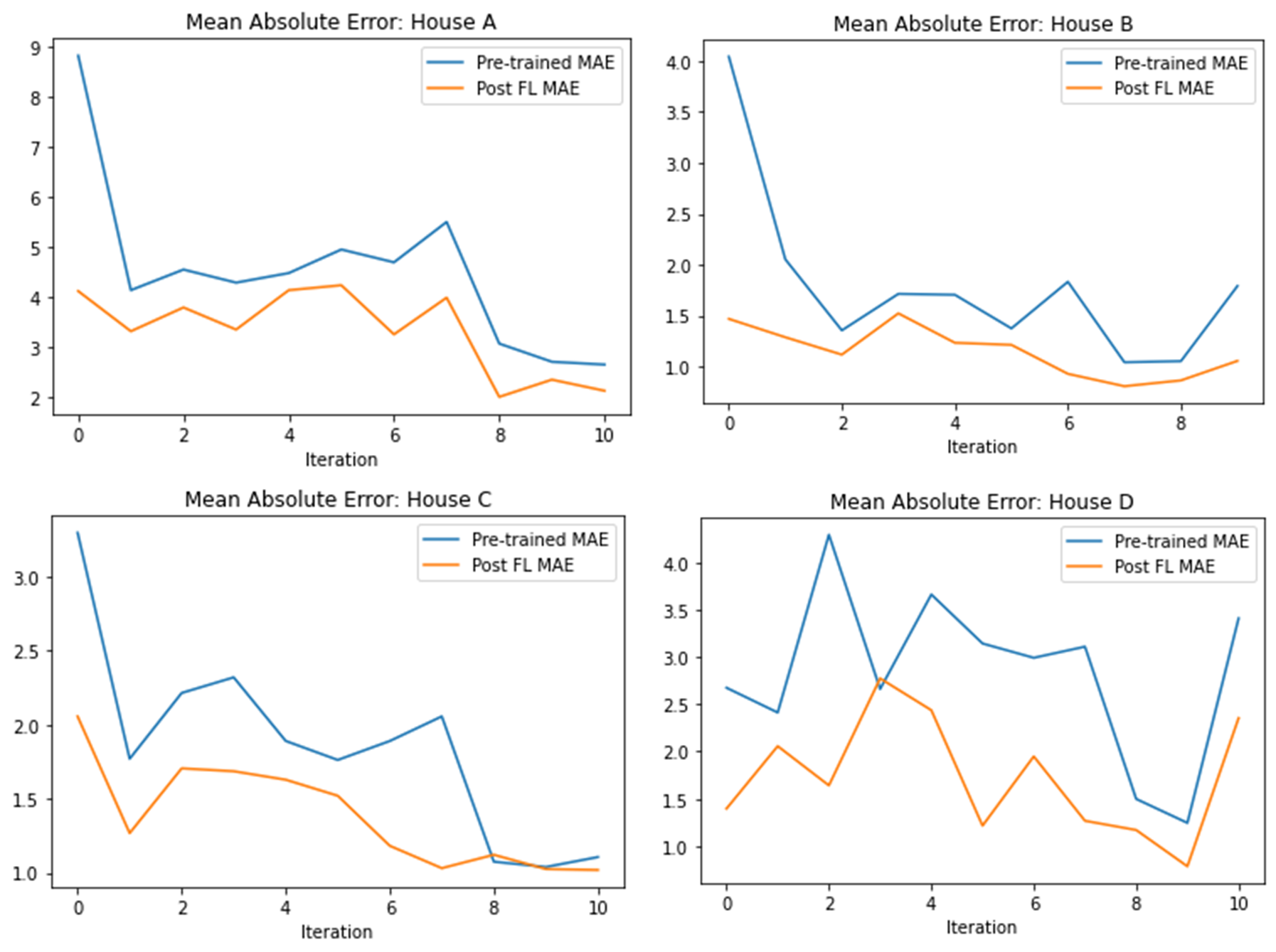}
    \caption{MAE per 2-week iteration for all 4 houses (6-hr ahead)}
    \label{fig:maesix}
\end{figure}

6-hour ahead predictions: Figure \ref{fig:maesix}

\section{Conclusion}
The results demonstrate that federated learning is useful for the task of predicting energy consumption and solar power output in a network. Solar power output predictions in particular have good accuracy. 

To improve the overall accuracy of the energy predictions, we need to train the model with more data over a longer period. For example, energy consumption will likely have seasonal variations but our model is only trained on 6 months of data so is not able to account for these patterns. 

Furthermore, increasing the model complexity, by adding more layers or adding LSTM layers, for example, will also help improve prediction accuracy. We limited the size of our model due to resource constraints on the Upboard but did not run into issues with inference or training so can experiment with larger model sizes. 

In addition, having more nodes in the network will also help to improve the benefit that federated learning has on the updates.

\section{Future Work}
To extend this work further, we can change the model to predict day-ahead energy consumption and solar output forecasts (meaning the model would have 48 outputs instead of 2). This would require a more complex model and more training data to achieve good performance. Though having a few hours ahead predictions are helpful for optimal load control and demand response, day-ahead forecasts will provide more valuable information to act upon.

We also want to test our system on out-of-distribution data (e.g different city or usage patterns) to evaluate the robustness of the model and federated learning algorithm. 

To improve cost efficiency, we plan on potentially updating the Upboard hardware to raspberry pi zeros instead. We also believe that we have designed our application to be scalable, and hope to productize with additional features in the future. 

\bibliographystyle{plain}
\nocite{*}
\bibliography{references.bib}

\appendix
\section{Appendix}
\subsection{Code}
\href {https://github.com/Sanjana-Sarda/Energy_Prediction}{Github Repository}

\end{document}